\documentclass[11pt]{article}
\usepackage{amsfonts}
\usepackage{amssymb}
\usepackage{graphicx}
\usepackage{setspace}
\usepackage[round]{natbib}
\usepackage{amsmath}%
\usepackage{amsthm}%
\usepackage{bbm}
\usepackage{palatino}
\usepackage{paralist}
\usepackage{multirow}
\usepackage{booktabs}
\usepackage{dsfont}
\usepackage{indentfirst}
\usepackage{enumerate}
\usepackage{enumitem}
\usepackage{longtable}
\usepackage{bm}
\usepackage{caption}
\usepackage{subcaption}
\usepackage{mwe}
\usepackage{float}
\usepackage{booktabs}
\usepackage{lscape}
\usepackage[hyperindex,breaklinks]{hyperref}
\hypersetup{colorlinks=true,       
	linkcolor=red,       
	citecolor=blue,        
	filecolor=magenta,      
	urlcolor=cyan           
}  
\usepackage{hyperref}

\usepackage{xcolor,csquotes,bm}

\newcommand{\bluecheck}{}%
\DeclareRobustCommand{\bluecheck}{%
  \tikz\fill[scale=0.4, color=blue]
  (0,.35) -- (.25,0) -- (1,.7) -- (.25,.15) -- cycle;%
}

\usepackage{palatino}
\usepackage{xskak}

\usepackage{tikz}
\usetikzlibrary{shapes.misc}
\newcommand{\mytimes}{ \tikz[baseline=-.55ex] \node [inner sep=0pt,cross out,draw,line width=1pt,minimum size=1ex] (a) {};}

\theoremstyle{definition}

\title{Karpov's Queen Sacrifices and AI}
\date{September 12, 2021}
\author{	
         \makebox[.2\linewidth]{Shiva Maharaj\footnote{Shiva Maharaj is founder and CEO of ChessEd. Nick Polson is Professor of Econometrics and Statistics at Booth School of Business.}}\\
	\textit{\small  ChessEd}
	\and
	\makebox[.2\linewidth]{Nick Polson}\\
	\textit{\small  Booth School of Business}\\
	\textit{\small  University of Chicago}\\
}

\graphicspath{{../fig/}{../wine/}}
	\usepackage{skak}
\usepackage{xskak}

\begin{document}

\maketitle


\emph{Combinations with a queen sacrifice are among the most striking and memorable --- Anatoly Karpov}

\vspace{0.15in}

\begin{abstract}
\noindent 
Anatoly Karpov's Queen  \makebox[1em]{\symqueen}   sacrifices are analyzed.  
Stockfish 14 NNUE---an AI chess engine---evaluates how efficient Karpov's sacrifices are. 
For comparative purposes, we  provide a dataset on Karpov's Rook  \makebox[1em]{\symrook} and Knight   \makebox[1em]{\symknight}   sacrifices to test whether Karpov achieves
a similar level of accuracy. Our study has implications for human-AI interaction and how humans can better understand the strategies employed by black-box AI algorithms.
Finally, we conclude with implications for  human study in. chess with computer engines. 

\bigskip
\noindent {\bf Key Words:}  AI, AlphaZero, LCZero, Bayes, Chess, Karpov, Neural Network, Reinforcement Learning, Stockfish 14 NNUE.
\end{abstract}

\newpage

\section{Introduction}
\vspace{0.1in}

\emph{Chess is not a game. Chess is a well-defined form of computation. You may not be able to work out the answers, but in theory, there must be
a solution, a right procedure in any position----John von Neumann}

\vspace{0.1in} 
The advent of computer chess engines based, such as AlphaZero, LCZero and Stockfish 14 NNUE, provides us with the ability to study optimal play.
AI chess algorithms are based on pattern matching, efficient search and data-centric methods rather than rules based.  Together with an objective functions based on maximising the probability of winning, we can now see what optimal  play and strategies look like. One caveat is the black-box nature of these algorithms and lack of  insight into the 
features that are empirically learned from self play. Therefore, we still need human intuition and explanation to fully understand  the principles and strategies
employed by these AI algorithms. Ai algorithms are based on Bellman's principle of optimality and the logic that  there is  the most precise way to continue.
Improvements are made only in search or computational complexity. Given the objective of maximising the chances of winning there is an optimal policy  to find from any given position.

Our goal is to show that  the play of Anatoly Karpov---the 12th world chess champion---is a natural experiment that mimics the play of AI chess engines.
Are Karpov's sacrifices sound? Section 1.1 provides a full description. and we will make comparisons with the strategies used by modern chess engines. 
On the empirical study, we study two datasets. First, we analyze a dataset of Karpov's $16$ known Queen sacrifices. Second, we analyze a dataset of  equivalent size on rook and knight sacrifices to see if there are any significant differences in optimality. 
Stockfish 14 NNUE measures the optimality of such moves in terms of centi-pawn loss. 

Sacrifices are different from Gambits---which are sub-optimal----in that they don't necessarily give up any advantage in the current  position. 
We find that Karpov's sacrifices are optimum more than 90\% of the time. A remarkable high percentage. 
We tend to associated sacrifices with being human and made for other reasons.  For example, the world-class chess players 
Tal and Nezmdhetdinov where famous for unsound sacrifices. Tal's logic (contrary to that of a rational chess engine) was the following:
\emph{You must take your opponent into a deep dark forest where $2+2=5$, and the path leading out is only wide enough for one}.

The rest of the paper is outlined as follows. Section 1.1 describes Karpov's approach to chess.. Karpov was unique in his "style" of play 
and we make comparisons to strategies employed by modern day chess engines. 
Section 2 provides a brief history of computer chess. A description of the AI underlying the applications of deep neural networks for chess engines is provided.
We link such optimal play with that of Karpov's.
Section 3 analysis Karpov's 16 known queen sacrifices to assess their optimality.  We find nearly all of his moves are optimal with an even higher performance for rook and knight sacrifices. For comparative purposes, we analyze a similar set of knight and rook sacrifices to see if there is any difference in optimality.   Specific board positions are analyzed to illustrate the Queen sacrifices. We also analyse  his immortal game versus Veselin Topolov where Karpov sacrificed a knight and then a rook for a bishop before winning.
Finally, Section 4  concludes with directions for future research.

\subsection{Karpov's "Style" of Play}

\vspace{0.1in}
\emph{Style?  I have no style. I play the position from where it is----Karpov on Karpov}

\vspace{0.1in}

Anatoly Karpov emerged as a challenger to world champion Bobby Fischer in the early 1970s. Karpov was declared World chess champion because bobby Fischer refused to play against him. It is the belief of many that Bobby Fischer made unreasonable demands for the match.  Fischer had a ruthless, aggressive and very precise form of play. 
Anatoly Karpov games showed that he played super-solid non-risk chess.

As evidence of this he tended to play \makebox[2em]{\symbishop e2}  against the Sicilian Najdorf also known as the rock. 
Karpov was described as begin the master of prophylactic play. It was once joking said that Karpov was able to snuff out ideas before they even entered your head. There was a lot of criticism after he won the world championship by default. Karpov then went on for the next ten years to play in every single major chess tournament  worldwide coming first or joint first in all of them,. He then said "I am second to none in chess". 

Karpov's style of play has been widely discussed and various attempts have been made to describe his style. It's a difficult undertaking to describe a specific style in chess since Karpov despite being seen as a highly positional chess player demonstrated uncanny very volatility tactical genius when the position called for it.  For the most part, analysts claimed that Karpov would build up small almost microscopic advantages and then the opponents position would fall apart. There was a joke that people lost to Karpov you do nothing wrong but you lose.  His ability to formulate highly intriguing plans made him an exceptionally and worthy chess champion.  Restricting your opponents counter play is key.
 In chess it is said that one should plan the opening as a book, the middle game as a magician, the endgame as a machine. Karpov demonstrated his mastery in all three fields. 

\begin{figure}[H]
\begin{center}
		\includegraphics[width=0.95\linewidth]{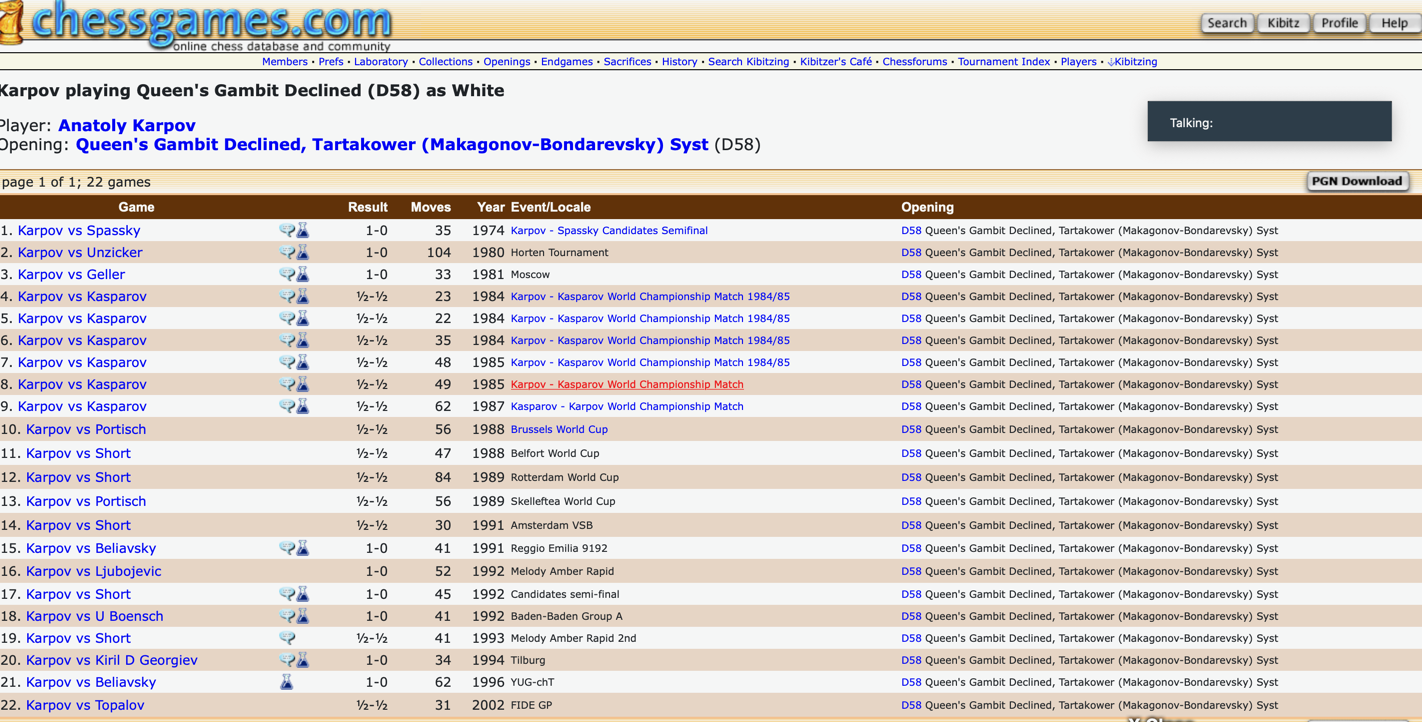}
\end{center}
\caption{Karpov's Queens Gambit}
\end{figure}

However, the question of style keeps coming up: \emph{What was his style?} His reputation for the most part has been that of a positional player.  In personal conversations, he did not seem to enjoy that description of himself and mentioned  that his games that numerous spectacular sacrifices.  Bobby Fischer said that tactics flow from a superior position.
So a very good understanding of placement of pieces on good squares, understanding pawn weaknesses, space advantages, intuitive and speed of attach together with king safety are 
necessary elements before tactical fireworks.  Karpov at one time said, "style, I have no style, I simply play the right moves". Karpov also said that it a position is draw, particularly  in the endgame, that no amount of will-power will change the result, "a draw is a draw". Judging from his openings he tented to play the Sicilian defense and the black-side of the Ruy Lopez.
As white is played both $e4$ and $d4$, making him a universal player.  Table 1 shows his incredible record with the Queens Gambit. 

A very beautiful game, that was played with computer-like precision was Anatoly Karpov vs Eldis Cobo Arteaga in 1972.  The precision of the attack and the control of space after freezing the central and queen-side of the board, led to a onslaught on the king-side with a deadly attack.  A rook sacrifice happened on move 29 with
\symrook xf6+. Computer chess engines amplify the traits that Karpov demonstrated in many cases with greater precision.  Anatoly Karpov is probably the closest human-being to play like modern-day computing machines and as such is worthy of detailed computer analysis of his games.

\section{Computer Chess: A Brief History}

John von Neumann described  chess as  a two-player zero-sum game with perfect information and proved the minimax theorem in 1928. His famous quote about chess being a form of computation has been realised in modern day AI with the use of deep neural networks to calculate $Q$-values in Bellman's equation.  This has a number of important ramifications for human play----in a given position there is an optimal action. There is no "style" of play---just optimal play where 
one simply optimises the probability of winning and follows the optimal Bellman path of play. The strategy of Exploitation and Exploration is central to dynamic programming.

Chess AI was pioneered by the three fathers of AI, namely Turing, Shannon and von Neumann.  Turing (1946) developed AI algorithms for chess playing. 
Shannon  (1950) described the how to program a computer to play chess and
von Neumann (1955) built the MANIAC computer and programmed it for chess where it took $12$ mins to search $4$ moves deep.
Modern day methods are based on $Q$ learning (a.k.a reinforcement learning).  These are NP-hard computational problems. The Shannon number which measures the number of possible board states is $ 10^{152} $ for chess making the computational challenge daunting. 

One approach are look-ahead calculation and search engines with complicated evaluation functions such as Stockfish 14 NNUE who use shallow networks for easy of evaluation. 
Another class of algorithms are deep neural networks, such as AlphaZero (Silver et al, 2017, Dean et al, 2012, Sadler and Regan, 2019) and LCZero, which are 
estimated from empirical self -lay and then used to interpolate the value and policy functions.  The goal is to simply maximize probability of winning.
The original chess engine algorithms used hand-coded rules and logistic regression from $23$ million games 
predicting the  next move of Grandmaster given the  board position.

Predicting when a computer would beat a human has an interesting history as well. 
Simon (1957) said it would be within $10$ years before it will beat a world champion. This didn't happened until 1992 with Kasparov playing deep blue. 
In 1958 the first human lost to computer lost to von Neumann's program. 
Botvinnik (1963) predicted that a Russian program will beat the World Champion 
In 1965 there was a famous game between Botvinnik and Shannon who travelled to Moscow solely to play the world champion. 

\subsection{NNs Chess Engines}

The dynamic programming method, known as $Q$-learning, breaks the decision problem into smaller sub-problems. Bellman's principle of optimality describes how to do this:

\vspace{0.1in}

\emph{Bellman Principle of Optimality: An optimal policy has the property that whatever the initial state and initial decision are, the remaining decisions must constitute an optimal policy with regard to the state resulting from the first decision. (Bellman, 1957)}

\vspace{0.1in}

Backwards Induction identifies what action would be most optimal at the last node in the decision tree (a.k.a. checkmate). Using this information, one can then determine what to do at the second-to-last time of decision. This process continues backwards until one has determined the best action for every possible situation (a.k.a solving the Bellman equation).

\vspace{0.1in}

\noindent{\bf Chess NNUEs}. First, one needs an objective function.  In the case of chess it is simply the probability of winning the game. Chess engines optimize the probability of a win via Bellman's equation and  use deep learning to evaluate the value and policy functions. 
The value function $V(s)$ is simply the probability of winning (100\% (a certain win) to 0\% (a certain loss). For a  given state of the board, denoted by $s$, the value function is given by 
$$
V(s) = P \left ( \textrm{winning} | s \right )  .
$$
The corresponding $Q$-value is probability of winning, given policy or move $a$, in the given state $s$ \textit{and} following the optimal Bellman path thereafter, we write 
$$
Q(s,a) = P \left ( \textrm{winning} | s , a \right ) 
$$
NN engines like AlphaZero don't use centi-pawn evaluations of a position but we can simply transform from centi-pawns to probabilities as follows:
The Win probability $P \left ( \textrm{winning} | s \right ) $ is related to centi-pawn advantage $c(s)$ in state $s$ of the board via the identity 
$$
w(s)= P \left ( \textrm{winning} | s \right )  = 1/ ( 1 + 10^{- c(s)/4 } )  \; \; {\rm and} \; \; c(s) = 4 \log_{10} \left ( w(s)/(1-w(s) )  \right ) 
$$
Hence this will  allow us to test the rationality of Gambits by measuring the difference between optimal play and gambit play using the optimal Bellman $Q$-values
weighted by the transition probabilities $ p( s^\star | s , a) $, estimated from human databases. 
At the beginning of the game, Stockfish 14 estimates that the centi-pawn advantage is  $ c(0) = 0.2 $ corresponding to $P \left ( \textrm{white winning}  \right )  = 0.524 $. 

The optimal sequential decision problem is solved by $Q$-learning (Polson and Sorensen, 2011, Polson and Witte, 2015) which calculates the $Q$-matrix, denotes by $Q(s,a)$ for state $s$ and action $a$. The goal is to maximise expected utility (von Neumann and Morgensterm 1944). 
The $Q$-value matrix describes the value of performing action $a$ (chess move( in our current state $s$ (chess board position) and then acting optimally henceforth.

The current optimal policy and value function are given by 
\begin{align*}
V(s) & = \max_a \; Q ( s , a )  = Q( s , a^\star (s)  )  \\
 a^\star (s) & = {\rm arg max}_a \;   Q ( s , a ) 
\end{align*} 
LCZero simply takes  the probability of winning as the objective function. Hence at each stage $V(s)$ measures the probability of winning.
This is typically reported as a centi-pawn advantage. 

The Bellman equation for $Q$-values becomes (assuming that the instantaneous utility $u(s,a)$) and that the $Q$ matrix is time inhomogeneous, is the constraint 
$$
Q( s , a) = u(s,a)+  \sum_{ s^\star \in S } P( s^\star | s ,a ) \max_{ a } Q ( s^\star , a )
$$
Here $P (s^\star | s ,a )$ denotes the transition matrix of states and describes the probability of moving to new state $ s^\star $ given current state $s$ and action $a$.
The new state is clearly dependent on the current action in chess and not a random assignment. 
Bellman's optimality principle is therefore simply describing the constraint for optimal play as one in which the current value is a sum over all future paths of the 
probabilistically weighted optimal future values of the next state.The right hand side  is a combination of reward and continuation value.

Taking maximum value over the current action $a$ yields 
\begin{align*}
V(s) &  =  \max_a  \left \{ u(s,a) +  \sum_{ s^\star \in S } P( s^\star | s ,a )   V (s^\star) \right \}  \; \; {\rm where} \; \; 
 V (s^\star)  = {\rm max}_a \;   Q ( s^\star , a ) .
\end{align*} 
Deep Neural Networks have achieved much success in learning how to play chess. By extracting nonlinear features from a large dataset of self-play games, algorithms 
such as AlphaZero or LCZero are able to estimate optimal policy and value functions required to maximize the probability of winning.  
How do deep neural networks work? Figure 1 shows how the policy and value functions reduce the dimensionality of the search by restricting breath and depth of the Monte Carlo 
Tres Search (MCTS). 
\begin{figure}
\includegraphics[width=1\linewidth]{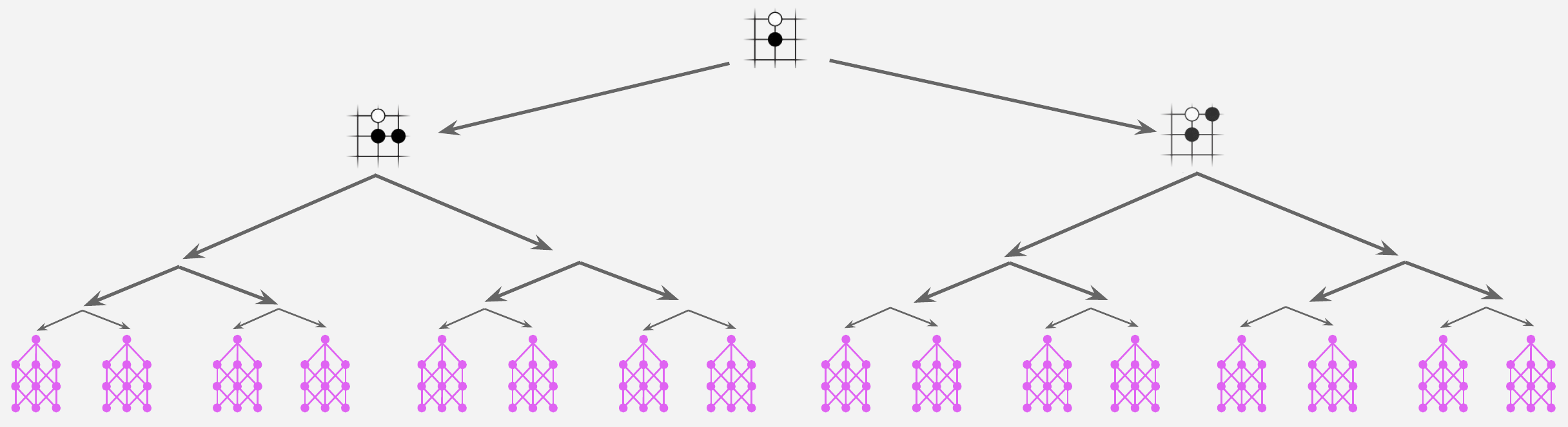}

\includegraphics[width=1\linewidth]{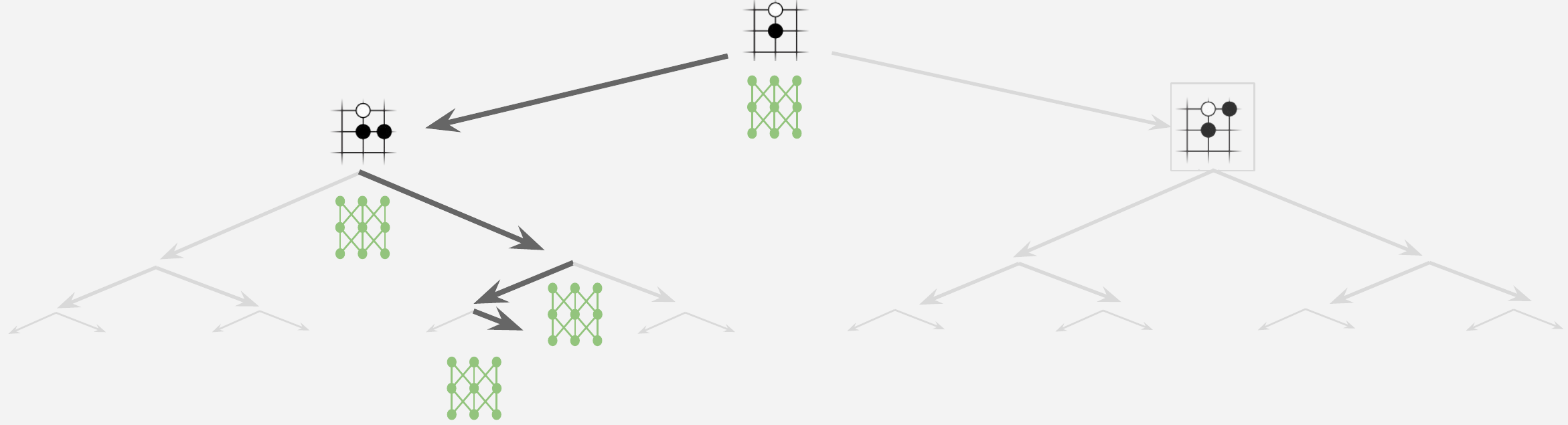}
\caption{Policy and Value Functions to reduce tree search. Source: Silver et al (2017)} 
\end{figure} 
This approach has the following three advantages 
\begin{enumerate}
	\item Value function approximates probability of winning. 
	\item Pick the path with highest approximated chance to win the game
	\item No need to explore the tree till the end
\end{enumerate}
We now turn to an analysis fo Karpov's Queen Sacrifices.

\section{Karpov's Sacrifices}

In this section, we consider two novel datasets. First, the dataset on the 16 known Karpov \symqueen sacrifices. Second, a dataset of 16  \symrook or \symknight
sacrifices to compare the accuracy of Karpov's play with regard to sacrifices. 

\subsection{Queen Sacrifices}

First, we analyze on his $16$ queen sacrifices\footnote{There is one other \symqueen  sac game. \href{https://lichess.org/UU6lG5xl}{Karpov vs
J. Polgar}. This is thought to be mysterious due to the nature of unsound queen sacrifice, The YouTuber Agadmator speculates that Karpov 
played the game blindfold. Karpov immediately resigned after the blunder.}.
Table 1 shows the $16$ Karpov Queen Sacrifices. The question is simply is it Optimal Play? We mark 
 $  \bluecheck $ optimal and $ \textcolor{red}{\mytimes} $ sub-optimal  as measured by the chess engine Stockfish 14. Rather remarkably over 90\% of the sacrifices where optimal play.
 Moreover, a number of the games when to endgame, so these sacrifices where not simple combinations of moves with an obvious checkmate on the horizon.

 For the sub-optimal games, we also provide the centi-pawn loss from optimality. Only one game had a material difference in the long run---the game with  Flores.
 Here Karpov's \symqueen sac cost him $2.1$ centi-pawns and the game. What was more interesting is the fact the Flores also sacked his \symqueen !
 Hence, this was a battle of extreme surprising tactics. 
\begin{table}[ht]
\centering 
\begin{tabular}{|c | c|} 
\hline	
\href{https://lichess.org/PfP7BaoG}{Karpov vs Timman} &  $ \textcolor{red}{\mytimes} $  $-1.1 $ \\
\href{https://lichess.org/vdsAKx53}{Karpov vs Ribli} & $\bluecheck $ \\
\href{https://lichess.org/y7IW9P5S}{Tatai vs Karpov}& $\bluecheck $ \\
\href{https://lichess.org/nSJjooS6}{Karpov vs Nedelin} & $\bluecheck$ \\
\href{https://lichess.org/uzOxZG0w}{Cordoba vs Karpov}& $\bluecheck$ \\
\href{https://lichess.org/5I2u20Dj}{Yakovich vs Karpov} &  $ \textcolor{red}{\mytimes} $  $-1.0 $ \\
\href{https://lichess.org/3WymLrly}{Anand vs Karpov} &  $\bluecheck $ \\
\href{https://www.chessgames.com/perl/chessgame?gid=1018838}{Karpov vs Anand} & $\bluecheck$ \\
\href{https://lichess.org/C5EJgum1}{Karpov vs Topalov}(Queens were traded)&  $\bluecheck $ \\
\href{https://lichess.org/lMO1qykc}{Karpov vs Gelfand} & $\bluecheck $ \\
\href{https://lichess.org/07PjVaPW}{Karpov vs Campora}(Both sides sacrificed their Q) & $\bluecheck $ \\
\href{https://lichess.org/ISlTcLWy}{Kurajica vs Karpov}(Both sides sacrificed their Q)&  $\bluecheck $ \\
\href{https://lichess.org/MWHXOcjy}{Karpov vs Adianto} &  $\bluecheck $ \\
\href{https://lichess.org/Ho2zfNYs}{Flores vs Karpov} &  $ \textcolor{red}{\mytimes} $  $-2.1 $ \\
\href{https://lichess.org/aUdZ4APF}{Ghaem Maghami vs Karpov} & $\bluecheck$ \\
\href{https://lichess.org/kj16eXtO}{Karpov vs Krysztofiak}&  $\bluecheck$\\  \hline
\end{tabular}
\caption{Queen Sacrifices;  $ \bluecheck $ optimal} 
\end{table} 

We now turn to specific games. The move sequence around the \symqueen sac are also given. 

\begin{center}
\fenboard{3rn1k1/5ppn/1p1P4/1r2pPP1/2q1P3/5BK1/1R5Q/3R4 w q - 0 1}
\showboard \\
\emph{Figure 1: Karpov vs. Ribli}
\end{center}
Move sequence:  \symqueen h7+  \symbishop xh7 \symrook h2+ \symking g8 \symrook dh1 f6 \symrook h8+. Ribli resign
\begin{center}
\fenboard{r3r1k1/1p4bp/6p1/8/1p1qp1b1/P5P1/1PQ1PPBP/R2NK2R b KQq - 0 1}
\showboard \\
\emph{Figure 2: Tatai vs. Karpov}
\end{center}
Move sequence: \symqueen d3 exd3 exd3+ \symking d2 \symrook e2+  (checkmate move 30) 
\begin{center}
\fenboard{7Q/5kpp/5n2/4n1B1/4q3/5R2/PP4KP/R7 w - - 0 1}
\showboard \\
\emph{Figure 3: Karpov vs. Anand}
\end{center}
Move sequence: \symqueen xg7+ \symking xg7 \symbishop xf6+ \symking g6 \symbishop xe5 (went to endgame)

\begin{center}
\fenboard{2kr1b1r/1pp2ppp/p1P1p3/P3q3/1n6/2N1BB2/1P3PPP/R2Q1RK1 b Qk - 0 1}
\showboard \\
\emph{Figure 4: Karpov vs. Timman}
\end{center}

Move sequence: dxc6 \symrook xd1 (\symqueen sac)  cxb7+  (went to endgame) 

\subsection{Rook and Knight Sacrifices}

 Second, a dataset of $16$ of his rook and knight sacrifices\footnote{The data is available from the well-know chess commentator KingCrusher on his Youtube channel. }. 
Table 2 provides the list of games and their centi-pawn deficit relative to Stockfish 14. This dataset provides a great comparison set of sacs  purpose, on Karpov's rook and knight sacrifices. A similar pattern emerges.  If anything Karpov is more efficient when sacrificing his generals (\symrook and \symknight). 
Again  $\bluecheck$ indicates an optimal move and $\textcolor{red}{\mytimes} $  a sub-optimal sacrifice. 

\begin{table}[ht]
\begin{tabular}{|c | c|} 
\hline	
\href{https://lichess.org/aDpwGujT}{Karpov vs Veselin Topalov} "Karpov's Immortal" (1994), Linares (N, R for B later)  & $\bluecheck $ \\
\href{https://lichess.org/iBDTMAvE}{Karpov vs Viktor Korchnoi}Candidates (1974), Moscow (P+R)  & $\bluecheck $ \\
\href{https://lichess.org/C5EJgum1}{Karpov vs Veselin Topalov (1994)}, Dos Hermanas (N+B) & $\bluecheck $ \\
\href{https://lichess.org/Bmu7xmiM}{Timman vs Karpov (1979)}, Montreal (B + N) & $\bluecheck $ \\
\href{https://lichess.org/8puqZrxa}{Karpov vs Boris Gulko (1996)}, Oropesa del Mar (R + N + R) & $\bluecheck $\\ 
\href{https://lichess.org/dxmFLb6G}{Karpov vs Evgeny Gik (1968)}, Moscow (R) & $\bluecheck $ \\
\href{https://lichess.org/R6OVhMh2}{Karpov vs Viktor Korchnoi (1971)}, Leningrad (R + R) & $\bluecheck $ \\
\href{https://lichess.org/Gy6jEE61}{Karpov vs Eldis Cobo Arteaga (1972)}, Skopje (R) & $\bluecheck $ \\
\href{https://lichess.org/UB6SBmgU}{Karpov vs Boris Spassky} 9th Soviet Match (1973), Moscow (R) & $\bluecheck $\\
\href{https://lichess.org/yfHyAUeJ}{Karpov vs Miguel A Quinteros} Leningrad Interzonal (1973) (R) & $ \textcolor{red}{\mytimes} $  $-0.1 $ \\
\href{https://lichess.org/sQUDi8Ep}{Karpov vs John Nunn (1982)} Kings, London (R?) & $\bluecheck $ \\
\href{https://lichess.org/ECcSN6AQ}{Seirawan vs Karpov (1982)} Hamburg (N+R) & $ \textcolor{red}{\mytimes} $  $-0.1 $ \\
\href{https://lichess.org/2HirQOej}{Karpov vs Gyula Sax: Linares (1983)}, Linares (N + R) & $\bluecheck $ \\
\href{https://lichess.org/PrBNVqUJ}{Timman vs Anatoly Karpov: Kings (1984)}, London (P+ R) & $\bluecheck $ \\
\href{https://lichess.org/3qhsHSoO}{Kasparov vs Anatoly Karpov (1987)} World Champ Seville (offered free R, declined) & $\bluecheck $ \\
\href{https://lichess.org/kRf9JRXf}{Karpov vs Vladimir P Malaniuk} 55th USSR Champ (1988) (R) & $\bluecheck $\\  \hline
\end{tabular}
\caption{Rook and Knight Sacrifices: $  \bluecheck $ optimal} 
\end{table} 
One game stands out and it is known as Karpov's immortal game.  The board position and move sequence are provided in Figure 5.
 The long sequence of moves eventually ended in checkmate on move 39. With the initial sacrifice started at move 19.

\begin{center}
\fenboard{rq3rk1/3bbp2/p1npp1p1/1p6/2P2P2/1NN3P1/PP1Q1PB1/R3R1K1 w Qq - 0 1}
\showboard \\
\emph{Figure 5: Karpov vs. Topolov: Immortal Game}
\end{center}

Karpov-Topolov Immortal: N sacrifice. Move sequence: \symking c5 dxc5 \symqueen xd7 (19). Mate move 39.

\section{Discussion}\label{sec:discussion}

AI now plays a central role in Human Knowledge acquisition. Kasparov (2017) provides an interesting discussion of the interplay between machine intelligence and human creativity.
Polson and Scott (2018) provide a framework for machines and  humans working together. With the advent of powerful chess engines that have the ability to 
calculate long sequences of optimal moves, we can now see what types of strategies the computer likes versus human play. Anatoly Karpov was uniques in that his
"style" was very similar to optimal moves generated by the computer.  This is borne out in our analysis of his Queen sacrifices.  Rather than speculative moves, Karpov simply found the optimal move (over $90$\% of the time). Analysing his style of play helps us understand the strategies found by pattern matching in and AI algorithm. 
There will always be a question of computation and the unreasonable effectiveness of data.  For example, simulating a 100 million games is exponentially small relative to the Shannon number of total possible combinations $ 10^{152} $.

Good (1977) summaries the issue of human knowledge and machine intelligence very well 

\textquote{\textit{It should now be clear that dynamic probability is fundamental for a theory of practical chess, and has wider applicability. Any such procedure, such as is definitely required in non-routing mathematical research, whether by human or by machines, \emph{must} make use of subgoals to fight the combinatorial explosion ... The combinatorial explosion is often mentioned as a reason for believing in the impracticability of machine intelligence, but if this argument held water it would also show that human
 intelligence is impossible.
Perhaps it is impossible for a human to be intelligent, but the real question is whether machines are necessarily equally unintelligent.} }

There are many outstanding problems. With improved computation and architectures for evaluating value and policy functions, it is possible that we will
find that centi-pawn advantages are actually larger than currently found. Moreover, many computer self-play games end in perpetual check, something that is rare with human play. 

\newpage
\section{References}
\noindent Bellman, R. (1957). \emph{Dynamic Programming}. Princeton University Press.\medskip 

\noindent Dean, J. et al (2012). Large Scale Distributed Deep Networks. \emph{Advances in Neural Information Processing Systems},  25, 1223-1231.\medskip

\noindent Good, I.J. (1977).  Dynamic Probability, Computer Chess, and the measurement of Knowledge. In: \emph{Machine Intelligence}.\medskip 

\noindent Karpov, A. (1992). \emph{Karpov on Karpov:  Memoirs of a Chess World Champion}. \medskip

\noindent Kasparov, G. (2017). \emph{Deep Thinking: when Machine Intelligence Ends and Human Creativity Begins}. Perseus.\medskip 

\noindent Silver, D. et al (2017). Mastering the Game of Go without Human Knowledge. \emph{Nature}, 550, 354-359.\medskip

\noindent Polson, N.G. and J. Scott (2018). \emph{AIQ}. St. Martin's Press. Macmillian.\medskip

\noindent Polson, N.G. and M. Sorensen (2011). A Simulation-based approach to Stochastic Dynamic Programming. \emph{Applied Stochastic Models},  27(2), 151-163. \medskip

\noindent Polson, N.G. and J. Witte (2015). A Bellman View of Jesse Livermore. \emph{Chance}, 28 (1), 27-31.\medskip

\noindent Sadler, M. and N. Regan (2019). \emph{Game Changer: AlphaZero's Groundbreaking Chess strategies and the Promise of AI.}. 
New in Chess.\medskip


\noindent Shannon, C. E. (1950). Programming a Computer to Play Chess. \emph{Phioisophical Magazine, 7 (41), 314}.\medskip 

\noindent von Neumann, J. and O. Morgenstern (1944). \emph{Theory of Games and Economic Behavior}. Princeton. \medskip

\end{document}